\documentclass[5p,twocolumn]{elsarticle}
\usepackage{amsmath,amsfonts}
\usepackage{algorithmic}
\usepackage{algorithm}
\usepackage{array}
\usepackage[caption=false,font=normalsize,labelfont=sf,textfont=sf]{subfig}
\usepackage{textcomp}
\usepackage{stfloats}
\usepackage{url}
\usepackage{verbatim}
\usepackage{graphicx}
\usepackage{booktabs}
\usepackage{gensymb}
\usepackage{multirow}
\usepackage{tikz}
\DeclareRobustCommand{\ConcatSymbol}{
  \tikz[baseline=(c.base)]{\node[draw,circle,thick,inner sep=0.4pt](c){C};}
}
\usepackage{breqn}
\usepackage{pgfplots}
\usepackage{amssymb}

\journal{Neurocomputing}

\begin{document}


\begin{frontmatter}

\title{LBTCap: A Lightweight Bilateral Transformer for Real-Time Remote Sensing Image Change Captioning}

\author[a]{Licheng Zhang\corref{cor1}}
\ead{licheng.zhang@student.unimelb.edu.au}
\author[b]{Siew-Kei Lam}
\ead{assklam@ntu.edu.sg}
\author[a]{Naveed Akhtar}
\ead{naveed.akhtar1@unimelb.edu.au}

\cortext[cor1]{Corresponding author}

\affiliation[a]{organization={School of Computing and Information Systems, The University of Melbourne},city={Melbourne},
            postcode={3010}, 
            state={Victoria},
            country={Australia}}

\affiliation[b]{organization={Nanyang Technological University},country={Singapore}}
\begin{abstract}
Remote sensing image change captioning (RSICC) generates natural-language descriptions of semantic changes between paired remote sensing images (RSIs), supporting applications such as urban planning, disaster response, and environmental monitoring. Although recent methods achieve strong captioning accuracy, most overlook computational efficiency and inference speed, which are essential for real-time deployment in practice. To this end, we propose LBTCap, a lightweight RSICC framework built on a bilateral Transformer that jointly models pre- and post-change features for efficient processing of paired RSIs. Specifically, we introduce a bilateral attention mechanism for paired inputs: the two temporal images are projected into separate queries and keys by the same query and key matrices shared across the two images, the value is formed from their concatenation, and the two resulting attention maps are combined by a learnable, structurally bilateral weighting instead of a fixed subtraction. This design keeps both temporal branches explicit while remaining compact, and, together with a truncated backbone and grouped-query attention, LBTCap uses only 39.99M parameters, of which the change-aware encoder accounts for just 2.78M. Extensive experiments on two public RSICC datasets show that LBTCap matches or closely approaches the accuracy of state-of-the-art methods while using far fewer parameters and running at markedly higher inference speed, with the benefit of the bilateral formulation most pronounced in the low-resource setting, demonstrating a favorable accuracy--efficiency trade-off for practical RSICC.
\end{abstract}


\end{frontmatter}



\section{Introduction}\label{sec:intro}
Remote sensing image change captioning (RSICC) is an emerging task with significant practical value, as it enables the automatic description of semantic changes between paired remote sensing images (RSIs) \cite{chouaf2021captioning}, which is critical for applications such as urban planning, disaster monitoring, and environmental assessment \cite{zou2025remote}. A comprehensive overview of remote-sensing spatiotemporal vision--language models can be found in~\cite{liu2025remote}. The main challenge of RSICC lies in accurately detecting and describing subtle or fine-grained changes between pre- and post-change RSIs, especially under variations in scale, lighting, season, and viewpoint, which can significantly degrade performance \cite{chang2023changes}. Although state-of-the-art approaches have demonstrated impressive performance, most methods largely overlook real-time requirements, which are crucial for time-sensitive and safety-critical applications. Consequently, there is a pressing need to develop RSICC models that simultaneously achieve real-time inference and competitive accuracy, bridging the gap between practical deployment and state-of-the-art performance.

\begin{figure}[!t]
    \centering
\resizebox{\columnwidth}{!}{
\begin{tikzpicture}
  \begin{axis}[
      width=8cm, height=7cm,
      xlabel={Parameters (M)},
      ylabel={BLEU-4 (\%)},
      xmin=0, xmax=650,
      ymin=62, ymax=69,
      grid=both,
      grid style={dashed, gray!30},
      legend style={at={(0.5,1.05)}, anchor=south, legend columns=4},
      legend cell align=left,
      clip=false,
      enlarge x limits=0.03,
      enlarge y limits=0.02
    ]

\addplot[only marks, mark=square*, blue, mark size=3pt]
coordinates {(39.99, 65.94)}
node[pos=1, anchor=west, xshift=2pt] {\scriptsize 65.94};
\addlegendentry{Ours}

\addplot[only marks, mark=triangle*, red, mark size=3pt]
coordinates {(647.00, 62.87)}
node[pos=1, anchor=east, xshift=-2pt] {\scriptsize 62.87};
\addlegendentry{SparseFocus}

\addplot[only marks, mark=o, green, mark size=3pt]
coordinates {(81.51, 62.77)}
node[pos=1, anchor=west, xshift=2pt] {\scriptsize 62.77};
\addlegendentry{RSICCformer}

\addplot[only marks, mark=diamond*, orange, mark size=3pt]
coordinates {(91.4, 67.93)}
node[pos=1, anchor=west, xshift=2pt] {\scriptsize 67.93};
\addlegendentry{TISDNet}

\addplot[only marks, mark=+, purple, mark size=3pt]
coordinates {(285.5, 64.39)}
node[pos=1, anchor=west, xshift=2pt] {\scriptsize 64.39};
\addlegendentry{Chg2Cap}

\addplot[only marks, mark=x, brown, mark size=3pt]
coordinates {(39.9, 64.09)}
node[pos=1, anchor=west, xshift=2pt] {\scriptsize 64.09};
\addlegendentry{SEN}

\addplot[only marks, mark=asterisk, cyan, mark size=3pt]
coordinates {(96.4, 66.12)}
node[pos=1, anchor=west, xshift=2pt] {\scriptsize 66.12};
\addlegendentry{ICT-Net}

\addplot[only marks, mark=star, magenta, mark size=3pt]
coordinates {(80.24, 65.41)}
node[pos=1, anchor=east, xshift=-2pt] {\scriptsize 65.41};
\addlegendentry{CD4C}

\addplot[only marks, mark=pentagon*, teal, mark size=3pt]
coordinates {(408.58, 63.54)}
node[pos=1, anchor=east, xshift=-2pt] {\scriptsize 63.54};
\addlegendentry{Prompt-CC}

\addplot[only marks, mark=triangle, pink, mark size=3pt]
coordinates {(309.55, 65.30)}
node[pos=1, anchor=east, xshift=-2pt] {\scriptsize 65.30};
\addlegendentry{KCFI}

\addplot[only marks, mark=diamond, violet, mark size=3pt]
coordinates {(176.90, 65.24)}
node[pos=1, anchor=west, xshift=2pt] {\scriptsize 65.24};
\addlegendentry{RSCaMa}

\addplot[only marks, mark=square, olive, mark size=3pt]
coordinates {(94.14, 64.62)}
node[pos=1, anchor=west, xshift=2pt] {\scriptsize 64.62};
\addlegendentry{TACC}

\addplot[only marks, mark=otimes, cyan!60!black, mark size=3pt]
coordinates {(243.19, 64.67)}
node[pos=1, anchor=west, xshift=2pt] {\scriptsize 64.67};
\addlegendentry{SFEN}

\addplot[only marks, mark=oplus, magenta!70!black, mark size=3pt]
coordinates {(73.0, 65.56)}
node[pos=1, anchor=west, xshift=2pt] {\scriptsize 65.56};
\addlegendentry{DACC}

\addplot[only marks, mark=Mercedes star, gray, mark size=3pt]
coordinates {(5.05, 64.38)}
node[pos=1, anchor=west, xshift=2pt] {\scriptsize 64.38};
\addlegendentry{Change3D}

\addplot[only marks, mark=star, black, mark size=3pt]
coordinates {(70.88, 68.52)}
node[pos=1, anchor=west, xshift=2pt] {\scriptsize 68.52};
\addlegendentry{DAE}

  \end{axis}
\end{tikzpicture}%
}
    \caption{Compared to state-of-the-art RSICC methods, our technique achieves competitive performance with a substantially reduced number of parameters.}
    \label{fig:acc_param}
\end{figure}
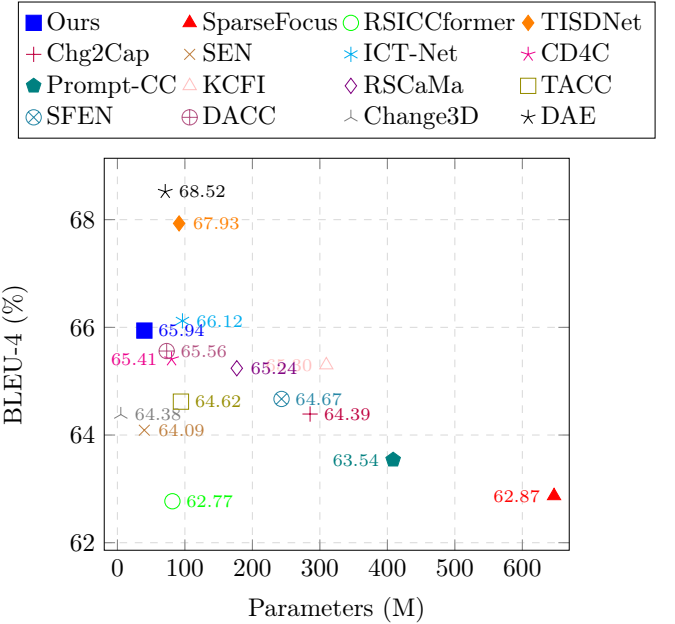

Prior research in RSICC can generally be divided into five categories. The first encompasses classical convolutional neural network (CNN) or recurrent neural network (RNN)-based approaches, as well as other traditional learning-based methods \cite{luppino2022code,ferrod2024towards,karimli2024data,sun2025scene,li2025cd4c,liu2024pixel,zou2025frequency,zhang2024scale,peng2024change,yang2025restricted}. The second employs diffusion-based generative models to iteratively map visual change representations into textual descriptions \cite{cai2024image,sun2025mask,yang2024remote,yu2025diffusion,bai2025cross}. Third, a range of approaches utilize large language models (LLMs) or vision--language models (VLMs) to perform change-aware captioning \cite{liu2024change,irvinteochat,bazi2024rs,yildirim2025llava,yang2025enhancing,deng2025changechat}. The fourth group uses Transformers or attention mechanisms to capture dependencies across temporal image pairs~\cite{liu2023progressive,xu2024mfrnet,chen2024multi,chang2023changes,hu2025sccnet,wu2025cross,liu2022remote,sun2024lightweight,zhou2024single,shi2024multi,hang2025text,karaca2025robust,li2024inter,cai2023interactive,li2024detection}. Finally, the fifth category includes state-space model (SSM) methods, particularly architectures based on Mamba~\cite{gu2024mamba}, which efficiently model long-range spatial-temporal correlations \cite{ma2025cross,liu2024rscama,ma2025ihm}.

Although these approaches have achieved notable performance, most are not optimized for real-time operation in practical scenarios. For example, LLM-based approaches impose substantial computational costs, limiting their applicability in time-sensitive tasks. Diffusion-based methods rely on multiple iterative steps, which significantly increases computation and prevents real-time deployment. Standard Transformers are also computationally expensive if used without architectural modifications. CNN- and RNN-based approaches, as well as attention-based methods, often struggle to capture fine-grained changes between paired RSIs. Mamba-based approaches are computationally efficient owing to their linear complexity. However, they currently struggle to achieve high efficiency and superior performance simultaneously.

Transformers have been widely used in RSICC and have demonstrated excellent performance \cite{liu2023progressive, sun2024lightweight, xu2024mfrnet, wu2025cross}. However, the vanilla Transformer remains computationally expensive, making real-time inference challenging. To address this, Ye et al.~\cite{yedifferential} recently proposed the Differential Transformer (DIFF Transformer). The DIFF Transformer achieves comparable language modeling performance while requiring only about two thirds of the model size or training tokens of a standard Transformer~\cite{yedifferential}. Long-sequence evaluations further show that DIFF Transformer can effectively exploit extended contextual information. Moreover, it provides clear advantages for LLMs, including improvements in key information retrieval, hallucination mitigation, in-context learning, and mathematical reasoning. 

However, both the standard Transformer and the DIFF Transformer can process only a single input at a time, limiting their ability to jointly model the relationships and changes between paired RSIs. Motivated by its efficiency, we take differential attention only as a starting point and propose a new bilateral attention mechanism tailored to the paired-input RSICC task. Our mechanism generates queries and keys separately from the two inputs and derives the value jointly from both. The concatenated value fuses information from both images, while the separate queries and keys yield one attention map per image, and the two maps are then combined to weight that fused value. The original DIFF Transformer uses a differential attention mechanism, in which attention scores are computed as the difference between two separate softmax attention maps. This mechanism can work effectively when processing a single input, as both attention maps originate from the same input. However, when applied to paired inputs, the fixed subtraction hard-wires one input as the subtracted term, biasing the interaction toward the other. To remove that bias, we recast the combination of the two attention maps into a structurally bilateral form, in which neither temporal input is fixed as the subtracted term, so a single learnable weight can emphasize either branch as the RSICC data require.

To summarize, we make the following contributions:
\begin{enumerate}
    \item[1)] We propose LBTCap, a lightweight Transformer-based RSICC framework aimed at efficient inference, addressing the limited attention paid to efficiency in existing methods.
    \item[2)] We propose a new bilateral attention mechanism for paired-input change modeling. It generates queries and keys separately from the two temporal images while forming the value from their concatenation, so that semantic changes are modeled explicitly across the two inputs, and it combines the two attention maps through a learnable bilateral weighting that treats the pre- and post-change images consistently. While the mechanism retains the efficiency of differential attention, its paired-input formulation, free of a hard-wired subtraction, is new. Our ablations show that the bilateral formulation performs on par with differential attention when training data are abundant, and clearly surpasses it in low-resource settings.
    \item[3)] We conduct extensive experiments on two public RSICC datasets. Compared with state-of-the-art methods, our approach attains accuracy competitive with the best-performing models while using far fewer parameters and running at markedly higher inference speed (see Fig.~\ref{fig:acc_param}). Comprehensive ablation studies further verify the effectiveness of each component, and qualitative results show that our method produces more accurate and descriptive change captions than existing approaches.
\end{enumerate}

The remainder of this paper is organized as follows. Section~\ref{sec:related} provides a review of related work in RSICC. Section~\ref{sec:method} presents the details of the proposed approach. Section~\ref{sec:experiments} describes the experimental setup and discusses the results. Finally, Section~\ref{sec:conclu} concludes the article. 
\section{Related Work}\label{sec:related}
Remote sensing image change captioning has attracted increasing attention in recent years. In this section, we organize the existing literature according to the underlying modeling paradigms. As most prior works employed Transformer or attention mechanisms in the decoder, we focus primarily on the feature extractor and encoder modules.\subsection{CNN- and Other Learning-based Methods}
Earlier RSICC approaches primarily relied on CNN--RNN or CNN--SVM pipelines \cite{chouaf2021captioning, hoxha2022change}. Domain-specific affinity matrices were introduced in \cite{luppino2022code}. Contrastive learning was adopted as the encoder in \cite{ferrod2024towards}, while extensive data augmentation strategies were explored in \cite{karimli2024data}. Self-supervised multifrequency representations were further used for disturbance-robust captioning in~\cite{zhao2026disturbance}. In addition, graph convolutional networks were employed in \cite{sun2025scene}, and graph-structural relationship modeling was further explored in~\cite{yang2026gscc}. CNN- and frequency-domain designs have also been investigated, such as MFG-Net~\cite{liu2026mfg}. More recently, \cite{zhu2025change3d} treated bi-temporal images as short video sequences, resulting in a lightweight model. However, the inserted perception frame increases the number of input frames processed by the video encoder, which adds to the inference cost. Moreover, its performance on benchmark datasets remained relatively limited. Multi-task learning for joint change detection and captioning was widely employed in \cite{zhu2024semantic, li2024detection, liu2024pixel, zou2025frequency, li2025cd4c}. A recent multi-task example, X-Change~\cite{tuzlupinar2026x}, jointly performs change captioning and spectral-index (NDVI/NDWI) segmentation through a shared spectral-aware encoder. Satellite image captioning was addressed by using only bi-temporal RSI training data to mitigate the scarcity of satellite captioning datasets \cite{peng2024change}. Zhang et al.~\cite{zhang2024scale} proposed a siamese architecture consisting of a backbone, a receptive field fusion block, and a global feature enhancement block. Similarly, Yang et al.~\cite{yang2025restricted} proposed a low-cost cascade information network, where updates were performed via a cascade linguistic module drawing on information theory. Whereas effective to an extent, traditional machine learning--based approaches are often unable to capture fine-grained changes, which limits their performance.
\subsection{Diffusion Model-based Approaches}
Cai et al.~\cite{cai2024image} proposed a text-guided diffusion model to generate synthetic RSICC datasets. Yang et al.~\cite{yang2024remote} introduced a diffusion feature extractor paired with an attention-based difference encoder to extract discriminative information. Similarly, Yu et al.~\cite{yu2025diffusion} proposed a probabilistic diffusion-based model by constructing a condition denoiser to map caption distribution to a
Gaussian distribution. Sun et al.~\cite{sun2025mask} proposed a shift toward data distribution learning using diffusion models, reinforced by frequency-domain noise filtering. Bai et al.~\cite{bai2025cross} employed manifold mapping to model illumination variations, followed by Bayesian diffusion to enhance the modeling of cross-temporal image changes, combined with a dual-layer multicoding module with a difference enhancement and dual-attention image-text captioning strategy. More recent diffusion-based methods aim to improve efficiency and guidance, such as the Karras diffusion framework KD-RSCC~\cite{yu2025kd} and differential-loss-guided generation~\cite{yang2026rscc}, as well as diffusion-based data augmentation~\cite{yu2025boosting}. Diffusion-based methods show improved performance over traditional techniques. However, they must rely on multiple iterative steps at the inference stage, which substantially increases computational cost and hinders real-time deployment. Moreover, their training is often computationally prohibitive.
\subsection{LLM-based Approaches}
LLM-based methods are gradually gaining traction for the RSICC task. 
Liu et al.~\cite{liu2023decoupling} fed multiple prompts and visual features from a feature-level encoder into a frozen LLM for language generation. Later, \cite{liu2024change} integrated a multilevel change interpretation model as the eyes and an LLM as the brain. Building on this line, Change-UP~\cite{yang2025change} unifies multilevel change interpretation with an LLM for joint change detection and captioning. Similarly, Wang et al.~\cite{wang2024chareption} designed a lightweight change adapter module, seamlessly integrated into both the vision backbone and the LLM, requiring minimal learnable parameters while optimally adjusting representations. Irvin et al.~\cite{irvinteochat} developed a new vision and language assistant called TEOChat that could engage in conversations about temporal sequences of earth observation data. Bazi \cite{bazi2024rs} developed RS-LLaVA, an improved version of popular LLaVA model, specifically adapted for RSIs through a low-rank adaptation approach. Yang et al.~\cite{yang2025enhancing} exploited the intrinsic knowledge of LLMs via visual instructions and improved the accuracy of change features using pixel-level change detection. Y\i{}ld\i{}r\i{}m et al.~\cite{yildirim2025llava} proposed LLaVa-RS, an LLM focused on remote images with features such as image captioning and change captioning. Deng et al.~\cite{deng2025changechat} introduced a VLM designed for interactive remote sensing change analysis, employing multimodal instruction tuning to address complex queries. More recent efforts continue in this direction with interactive, instruction-tuned, and retrieval-augmented vision--language models~\cite{deng2026deltavlm,noman2025cdchat,li2026btcchat,duman2025blip,wang2025ragcc,leon2026describing,atecs2025scalable,xue2026towards}. LLM-based approaches are constrained by large model size, high computational cost, long inference time, and reliance on large annotated datasets.
\begin{figure*}[t]
  \centering
   \includegraphics[width=\linewidth, trim={0 0 0 40}, clip]{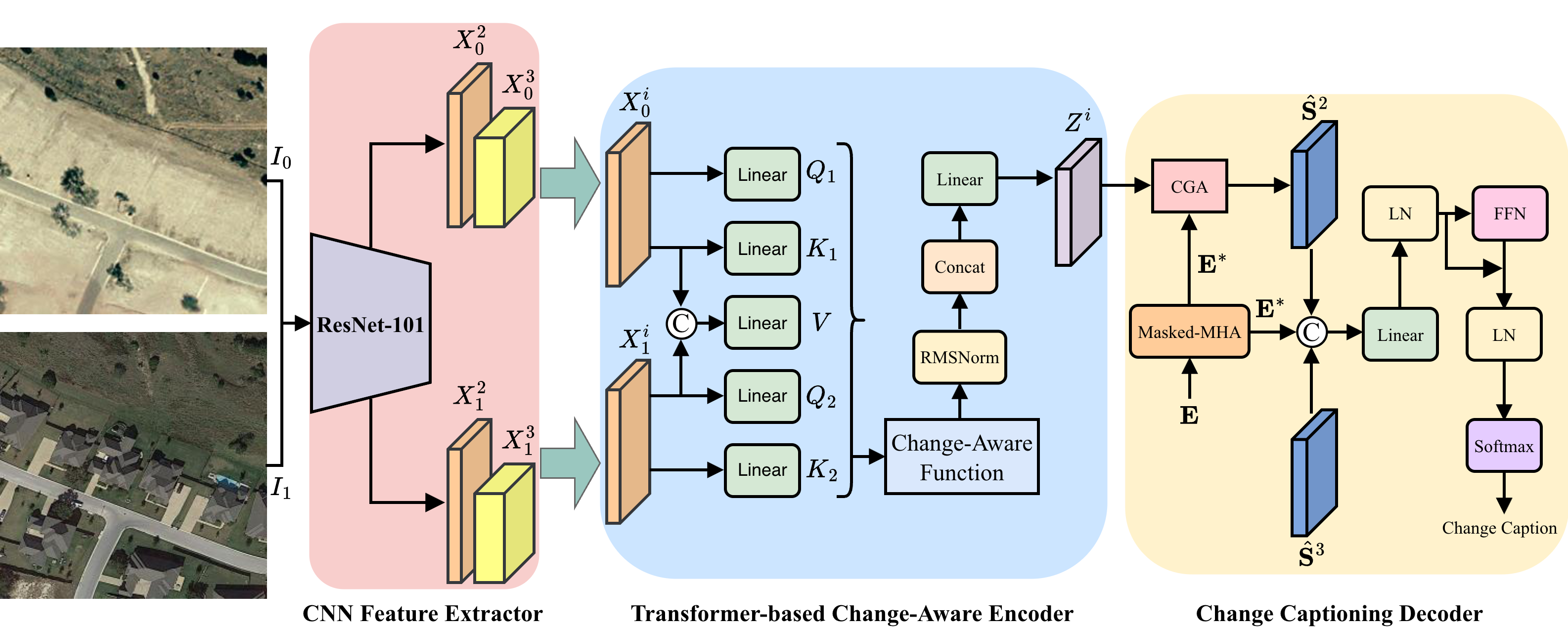}
   \caption{Overview of the proposed LBTCap. The backbone of ResNet-101 extracts multi-scale visual features. The change-aware encoder captures the semantic changes between each feature pair. The change captioning decoder generates change captions. \protect\ConcatSymbol~denotes concatenation along feature dimension.}
   \label{fig:LBTCap}
\end{figure*}
\subsection{Transformer and Attention-based Mechanisms}
In this category, pure Transformer architectures were proposed in \cite{liu2023progressive,wu2025cross}. Liu et al.~\cite{liu2023progressive} adopted ViT as the feature extractor and stacked multiple progressive difference perception layers to progressively exploit change features from bi-temporal inputs. Wu et al.~\cite{wu2025cross} employed SegFormer as the feature extractor and designed a cross-spatial Transformer and a symmetric differential localization module to capture changes. Many studies integrated CNN with attention \cite{xu2024mfrnet,chang2023changes} or Transformer \cite{sun2024lightweight,liu2022remote,shi2024multi,vyshnav2024intelli,hu2025sccnet}. Cross-attention was employed in \cite{li2024inter, zhou2024single, hang2025text} to capture interactions across two inputs. Zhou et al.~\cite{zhou2024single} proposed a lightweight model with a parameter count comparable to ours. However, it exhibited relatively slow inference and limited performance on benchmark datasets. Gated attention was used in \cite{cai2023interactive, chen2024multi, karaca2025robust}. A large body of recent attention-based methods further refine change modeling with difference-aware, cross-attention, and semantic-alignment designs~\cite{qin2025dgat,wang2025change,meng2025dualfocus,dang2026hcdfnet,li2026exploring,tang2025bitcc,huang2026dinov3,li2025salient,dong2025rscac,li2026multiscale,tang2025progressive,yang2026spatial,shi2026multi,sun2025global,xian2025dynamic,wang2025remote}, and several further pursue lightweight designs~\cite{sun2026scnet,gao2025swmcc,wang2025lightweight}. Standard Transformers remain computationally demanding, while attention-based methods often struggle to capture fine-grained changes because such changes intrinsically attract less attention. 
\subsection{Mamba-based Models}
In a recent stream of methods, Ma et al.~\cite{ma2025cross} combined Transformer and Mamba modules, employing cross-layer change awareness and temporal aggregation to capture bi-temporal correlations. Liu et al.~\cite{liu2024rscama} utilized spatial difference-aware and temporal-traversing state space models to refine bi-temporal features before language decoding. Ma et al. \cite{ma2025ihm} proposed an interactive hierarchical Mamba encoder with cross-fusion screening to select key change features for captioning. More recently, hybrid Mamba--Transformer designs such as MTH-Net~\cite{ma2025mth} and the CNN--VMamba synergistic network RMNet~\cite{cao2026rmnet} have been explored to combine linear-time modeling with stronger feature extraction. Although Mamba offers linear computational complexity, no existing model has yet achieved both high efficiency and high accuracy for the RSICC task.
\subsection{Summary and Positioning}
Across these paradigms, RSICC has evolved from early CNN--RNN pipelines toward Transformer- and attention-based encoders, and more recently toward LLM-, diffusion-, and Mamba-based models. The dominant trend has been to push captioning accuracy higher, typically at the cost of rapidly growing model size and inference latency: diffusion models require iterative sampling, LLM-based models carry billions of parameters, and even standard Transformer encoders are computationally heavy, while Mamba-based models improve efficiency but have not yet matched the accuracy of the strongest Transformer models. A second common thread is how change is modeled: most residual-plus-Transformer methods with multi-scale and cross-attention designs represent change by feature differencing or concatenation followed by attention over a single fused stream. Our work differs in the change-modeling mechanism rather than in the backbone: we propose a paired-input bilateral attention whose learnable weighting replaces the fixed subtraction, so that the two temporal inputs are treated consistently while the encoder remains lightweight. This positions LBTCap in the efficiency-oriented branch of RSICC, targeting a favorable accuracy--efficiency trade-off rather than maximal accuracy alone. In particular, recent lightweight or attention-based methods such as SCNet~\cite{sun2026scnet} and CASC~\cite{wang2025change} also build on residual backbones with multi-scale and cross-attention designs, yet they still model change through feature differencing or cross-attention over a fused stream. In contrast, our change-aware encoder keeps the two temporal inputs explicit and combines their two attention maps through a learnable bilateral weighting, which is the key distinction.
\section{Methodology}\label{sec:method}
In this section, we present LBTCap, a novel Transformer-based framework for RSICC, as illustrated in Fig.~\ref{fig:LBTCap}. The framework consists of three components: (1) a multi-scale feature extractor that obtains paired representations from multiple stages of a ResNet-101 backbone for bi-temporal RSIs, (2) a lightweight bilateral Transformer encoder that introduces a new paired-input bilateral attention to capture discriminative semantic changes from paired features, and (3) a multi-scale captioning decoder with cross-gated attention to fuse change-aware features for caption generation.
\subsection{Multi-Scale Feature Extraction}\label{sec3.1}
We employ a ResNet-101~\cite{he2016deep} backbone as the feature extractor to obtain multi-scale representations from bi-temporal RSIs. Given a pair of input images, denoted $I_0$ and $I_1$, the backbone encodes each into hierarchical feature pyramids. To ensure consistent dimensions across scales, each feature is projected into a shared embedding space via a learnable transformation, and a learnable 2D positional embedding is added at each scale to encode spatial information. To reduce model parameters, we discard the last stage of the backbone, retaining only the stem layers and stages 1--3, and take the outputs of stages 2 and 3 as the multi-scale features. This design reflects a deliberate trade-off between representational richness and computational efficiency. Features from stages 2 and 3 provide sufficient spatial resolution and strong semantic abstractions for capturing fine-grained changes, whereas stage 4 produces low-resolution representations that are computationally expensive but contribute little to change detection. The two multi-scale features for $I_0$ and $I_1$ are denoted as $X_0^2 \in \mathbb{R}^{2H \times 2W \times C}$ and $X_0^3 \in \mathbb{R}^{H \times W \times C}$ for $I_0$, and $X_1^2 \in \mathbb{R}^{2H \times 2W \times C}$ and $X_1^3 \in \mathbb{R}^{H \times W \times C}$ for $I_1$, respectively, where $H$ and $W$ refer to the height and width of the stage 3 feature maps, and $C$ denotes the unified channel dimension after transformation. We adopt ResNet-101 as the backbone to stay consistent with the majority of prior RSICC methods~\cite{cai2023interactive,li2024inter,liu2022remote}, so that our experiments isolate the effect of the proposed change-aware encoder rather than that of the backbone. Since our contribution lies in the encoder, it is backbone-agnostic and transfers directly to intrinsically lightweight backbones.
\subsection{Transformer-based Change-Aware Encoder}\label{sec3.2}
The key to RSICC lies in capturing changes between bi-temporal RSIs. However, a CNN backbone extracts features from each input independently. To this end, we propose a Transformer-based change-aware encoder that explicitly captures the changes between paired features. The proposed bilateral attention introduces four new design elements relative to standard differential attention~\cite{yedifferential}, which together constitute the novelty of this component. First, whereas the DIFF Transformer processes a \emph{single} input, our dual-input mapping generates queries and keys separately from the two temporal images, so that changes are modeled \emph{across} images rather than within a single one. Second, we share the query and key projections between the two inputs in a Siamese manner, placing both branches in a common space and tying them at the projection level. Third, we replace the subtractive combination of the two attention maps, which is prone to favoring one input, with a structurally bilateral weighted combination that treats the two inputs consistently. Fourth, the value is derived from the concatenation of both inputs, so that the fused representation carries joint bi-temporal information. Compared with prior residual-plus-Transformer RSICC encoders that model change by feature differencing or concatenation followed by attention over a single fused stream, our design keeps the two inputs explicit throughout the attention computation while remaining lightweight, which enables the accuracy--efficiency trade-off targeted in this work. We deliberately model change without an explicit feature-differencing step: because the value at every position is a joint bi-temporal descriptor, combining the two branch-specific attention maps yields, for each spatial location, a fused representation that carries both temporal views, and the learnable weight controls how strongly each view contributes. The change signal is therefore preserved in the encoder output and read out by the cross-gated decoder, rather than being collapsed early by a hard difference operation as in prior difference-based encoders, which can discard bi-temporal information before the decoder sees it.

\begin{figure}[t]
  \centering
   \includegraphics[width=\linewidth, trim={0 20 0 50}, clip]{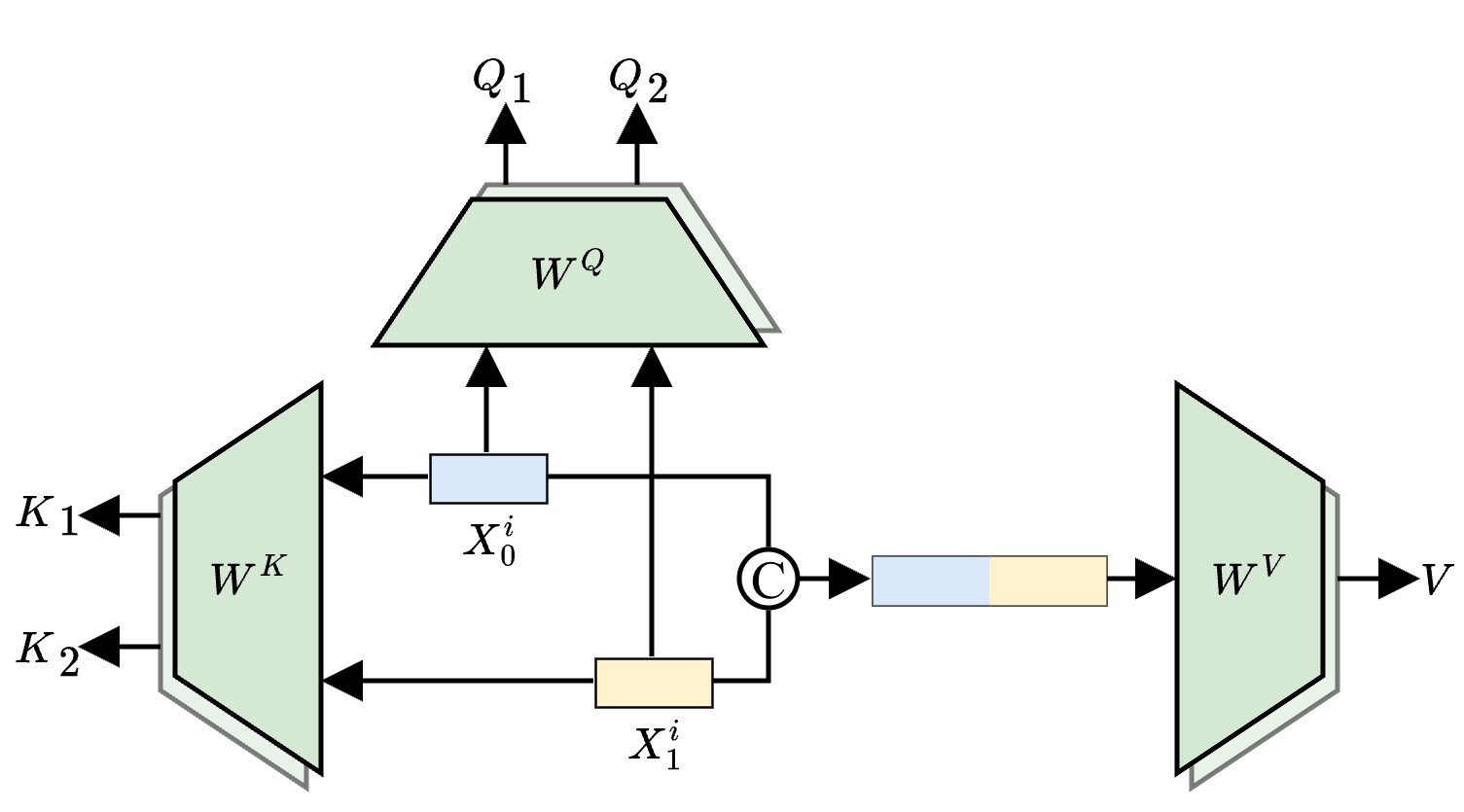}
   \caption{Illustration of the proposed query--key--value mapping method. $X_0^i$ and $X_1^i$ represent paired features.}
   \label{fig:qkv}
\end{figure}

\subsubsection{Dual-Input QKV Mapping}

We now detail the proposed bilateral attention. Whereas the DIFF Transformer~\cite{yedifferential} processes a single input and generates two attention scores from it, our encoder processes both temporal inputs simultaneously to capture changes between paired features. Specifically, $Q_1$ and $Q_2$, as well as $K_1$ and $K_2$, are computed separately from the two inputs, while the shared $V$ is constructed from the concatenated features of both inputs. 

Given a pair of feature representations \(X_0^i, X_1^i \in \mathbb{R}^{N \times C}\), with \(i \in \{2,3\}\), where \(N\) denotes the sequence length, the queries, keys, and value are computed as:
\begin{equation}
\begin{aligned}
Q_1 &= X_0^i W^Q, & K_1 &= X_0^i W^K, \\
Q_2 &= X_1^i W^Q, & K_2 &= X_1^i W^K, \\
V &= \text{Concat}(X_0^i, X_1^i) W^V,
\end{aligned}
\label{eqn:qkv}
\end{equation}
where $W^Q, W^K \in \mathbb{R}^{C\times d}$ and $W^V \in \mathbb{R}^{2C\times d}$ are learnable projection matrices, $d$ is the embedding dimension, and $\text{Concat}(\cdot,\cdot)$ concatenates the two features along the channel dimension. Fig.~\ref{fig:qkv} illustrates the generation of the key components used in the subsequent change-aware function computation. Notably, the two inputs share the same projection matrices in a Siamese manner: $Q_1$ and $Q_2$ are produced by the same $W^Q$, and $K_1$ and $K_2$ by the same $W^K$, rather than by separate per-input projections. This weight sharing keeps the two branches in a common embedding space, ties the pre- and post-change features at the projection level, and reduces the number of parameters. In contrast, $W^V$ operates on the concatenation of both inputs, so that the shared value carries information from the two images jointly.
\subsubsection{Change-Aware Function}
After $Q_1, Q_2, K_1, K_2$, and $V$ are derived, we compute the attention scores for the two inputs using $(Q_1, K_1)$ and $(Q_2, K_2)$, respectively. The resulting attention scores are then employed to capture the changes between paired features. To this end, we propose a bilateral change-aware function, which is mathematically defined as follows:
\begin{equation}
\begin{split}
\mathrm{Attn}(Q_1, K_1, Q_2, K_2, V) ={}& \Big( (1-\lambda)\,\mathrm{softmax}\big(\tfrac{Q_1K_1^{T}}{\sqrt{d}}\big) \\
& + \lambda\,\mathrm{softmax}\big(\tfrac{Q_2K_2^{T}}{\sqrt{d}}\big) \Big)V,
\end{split}
\end{equation}
where $\lambda$ is a learnable balancing weight
and $d$ denotes the dimension of the key.
Unlike~\cite{yedifferential}, which models change as the \emph{difference} between the first and second attention maps and therefore fixes the first map at a positive weight while subtracting the second, our function combines the two maps additively, so that both inputs enter the output through the same softmax-and-weighting operation. We use ``bilateral'' in a strictly structural sense: neither temporal branch is hard-wired as the added or the subtracted term, so the operator treats the two inputs interchangeably at the architectural level. Being bilateral in structure does not mean equal weighting. On the contrary, it is exactly what frees the single learnable scalar $\lambda$ to move emphasis onto either branch, and even to reverse which branch is suppressed, as the data require. The subtractive form of~\cite{yedifferential} lacks this property because it can only ever hold the first map positive and subtract the second. Here $\lambda$ is a single per-layer value. Following the DIFF Transformer~\cite{yedifferential}, it is not a raw parameter but is reparameterized exactly as in the original DIFF Transformer~\cite{yedifferential}, namely $\lambda = \exp(\boldsymbol{\lambda}_{Q_1}\!\cdot\!\boldsymbol{\lambda}_{K_1}) - \exp(\boldsymbol{\lambda}_{Q_2}\!\cdot\!\boldsymbol{\lambda}_{K_2}) + \lambda_{\mathrm{init}}$, where $\boldsymbol{\lambda}_{Q_1},\boldsymbol{\lambda}_{K_1},\boldsymbol{\lambda}_{Q_2},\boldsymbol{\lambda}_{K_2}$ are separate learnable vectors named consistently with the queries and keys in Eq.~\eqref{eqn:qkv}. We keep this reparameterization identical to~\cite{yedifferential}, and $\lambda$ is fixed after training. Importantly, $\lambda$ is unconstrained and may lie outside $[0,1]$, so that either the pre- or the post-change branch can be emphasized or fully suppressed. We keep $\lambda$ shared within a layer rather than conditioning it on each input, as this adds negligible parameters and, on the small RSICC training sets, avoids the extra variance of predicting a per-sample weight.
\subsubsection{Multi-Head Attention}\label{sec:gqa}
In line with standard Transformer practice, our Transformer employs multi-head attention (MHA). We adopt a Grouped Query Attention (GQA) mechanism, where multiple query heads share the same key and value matrices. In our implementation, we configure eight query heads and four key--value heads, so that each pair of query heads shares a single set of key and value projections. This design further reduces computational overhead.

Formally, MHA in our Transformer is formulated as:
\begin{equation}
\begin{gathered}
\text{head}_j = \text{Attn}\big(X_0^i, X_1^i; W_j^Q, W_{g(j)}^K, W_{g(j)}^V, \lambda\big), \\
\widehat{\text{head}}_j = (1 - \lambda_{\text{init}}) \text{LN}_{\text{RMS}}(\text{head}_j), \\
\text{MHA}(X_0^i, X_1^i) = \text{Concat}\big(\widehat{\text{head}}_1, \ldots, \widehat{\text{head}}_h\big) W^O,
\end{gathered}
\label{eqn:gqa}
\end{equation}
where $W_j^Q$, $W_{g(j)}^K$, and $W_{g(j)}^V$ are the learnable projection matrices for the $j$-th head, and $g(j)$ indicates the group index of the query head. Here $\text{Attn}(X_0^i, X_1^i; \cdot)$ is a shorthand for the change-aware function above, with the queries, keys, and value computed from $X_0^i$ and $X_1^i$ through the per-head projections as in Eq.~\eqref{eqn:qkv}. $\text{Concat}(\cdot)$ denotes concatenation along the feature dimension, and $W^O$ is the learnable output projection that aggregates the $h$ heads. The encoder normalization, denoted $\text{LN}_{\text{RMS}}$, is implemented via RMSNorm~\cite{zhang2019root}, to distinguish it from the standard layer normalization (LN) used in the decoder. $\lambda_{\text{init}}$ is parameterized as defined in~\cite{yedifferential}.
\subsubsection{Complexity}
For a bi-temporal pair with sequence length $N$ and embedding dimension $d$, the bilateral attention computes two softmax maps, $Q_1K_1^{\top}$ and $Q_2K_2^{\top}$, each of size $N\times N$ at cost $\mathcal{O}(N^{2}d)$, forms the shared value at cost $\mathcal{O}(NCd)$, and applies the combined map to the value at cost $\mathcal{O}(N^{2}d)$. The overall cost is thus $\mathcal{O}(N^{2}d)$, the same asymptotic order as a single-input self-attention and as the original differential attention, since handling the two temporal inputs adds only a constant factor rather than changing the complexity class. The memory footprint follows the same $\mathcal{O}(N^{2})$ attention cost, which grouped-query attention further reduces by sharing key and value projections across query heads. The paired-input design therefore incurs no asymptotic overhead over a standard lightweight Transformer encoder, consistent with the small parameter count (2.78M) reported in Section~\ref{sec:experiments}.
\subsection{Change Captioning Decoder}\label{sec3.3}
We then employ a change captioning decoder to generate change captions by drawing on the aforementioned multi-scale change-aware features, denoted as $Z^j$ with $j \in \{2,3\}$ indexing the feature scale. 
The decoder follows the design principle of~\cite{cai2023interactive}. It consists of three main submodules: masked MHA (Masked-MHA), cross gated-attention (CGA), and a feed-forward network (FFN). Each submodule is equipped with a residual connection and LN. Given a sequence of word embeddings $\mathbf{E} = \{\mathbf{E}_1, \dots, \mathbf{E}_L\} \in \mathbb{R}^{L \times D}$, where $L$ denotes the length of the sentence and $D$ is the word embedding dimension, Masked-MHA masks the subsequent positions at time step $t$ and learns to predict the word features
\begin{equation}
\begin{aligned}
\mathbf{E}^{*} &= \text{Concat}_{i=1}^h(\text{head}_i) W_o,
\end{aligned}
\end{equation}
where
\begin{equation}
\begin{aligned}
\text{head}_i &= \text{Masked\text{-}Attention}(\mathbf{E} S_i^Q, \mathbf{E} S_i^K, \mathbf{E} S_i^V),
\end{aligned}
\end{equation}
where $S_i^Q$, $S_i^K$, and $S_i^V$ are learnable projection matrices, and $W_o$ projects the concatenated outputs of the $h$ attention heads. $\text{Masked-Attention}(\cdot,\cdot,\cdot)$ denotes the masked attention operator.

Then, the CGA module performs cross-attention to integrate $\mathbf{E}^{*}$ with the multi-scale features:
\begin{equation}
\begin{aligned}
\mathbf{S}^j &= \text{Concat}_{i=1}^h(\text{head}_i) P_o,
\end{aligned}
\end{equation}
where
\begin{equation}
\begin{aligned}
\text{head}_i &= \text{Cross-Attention}(\mathbf{E}^{*} P_i^Q, Z^j P_i^K, Z^j P_i^V),
\end{aligned}
\end{equation}
where $P_i^Q$, $P_i^K$, $P_i^V$ are learnable projection matrices, and $P_o$ is the output projection matrix.

Then, gated attention is used to focus on relevant changes of interest, which is computed as:
\begin{equation}
\begin{aligned}
\hat{\mathbf{S}}^j &= \mathbf{g}_s^j \odot \mathbf{S}^j, \\
\mathbf{g}_s^j &= \sigma(\text{Linear}(\text{Concat}(\mathbf{E}^{*}, \mathbf{S}^j))),
\end{aligned}
\end{equation}
where $\sigma$ denotes the sigmoid function and $\odot$ is element-wise multiplication.

Finally, the multi-scale sentence features $\hat{\mathbf{S}}^j \in \mathbb{R}^{L \times D}$ are summed together:
\begin{equation}
\begin{aligned}
\mathbf{C} &= \text{LN}(\text{Linear}(\text{Concat}(\hat{\mathbf{S}}^2, \hat{\mathbf{S}}^3, \mathbf{E}^{*}))), \\
\mathbf{C} &= \text{LN}(\mathbf{C} + \text{FFN}(\mathbf{C})),
\end{aligned}
\end{equation}
where LN is realized with common layer normalization.

The output of the captioning decoder $\mathbf{C} \in \mathbb{R}^{L \times D}$ is then fed into a linear projection layer followed by a softmax layer to predict word probabilities over the vocabulary:
\begin{equation}
\mathbf{P} = \text{Softmax}(\text{Linear}(\mathbf{C})).
\end{equation}

The overall workflow of the proposed framework is summarized in Algorithm~\ref{alg:lstcap}.
\begin{algorithm}[t]
\caption{The LBTCap model}
\label{alg:lstcap}
\hspace*{\algorithmicindent}\textbf{Input:} Pre- and post-change RSIs $I_0$ and $I_1$, and word embeddings $\mathbf{E}$\\
\hspace*{\algorithmicindent}\textbf{Output:} Change caption
\begin{algorithmic}[1]
\STATE // Step 1: Multi-scale feature extraction
\FOR{$i \in \{0, 1\}$}
    \STATE $X_i^2, X_i^3 \leftarrow$ ResNet-101($I_i$)
\ENDFOR
\STATE // Step 2: Change-aware encoder
\FOR{$i \in \{2, 3\}$}
\STATE $Z^i \leftarrow$ MHA($X_0^i, X_1^i$)
\ENDFOR
\STATE // Step 3: Change captioning decoder
\STATE $\mathbf{E}^* \leftarrow$ Masked-MHA($\mathbf{E}$)
\FOR{$j \in \{2, 3\}$}
\STATE $\hat{\mathbf{S}}^j \leftarrow$ CGA($\mathbf{E}^*, Z^j$)
\ENDFOR
\STATE $\mathbf{C} \leftarrow$ LN(Linear(Concat($\hat{\mathbf{S}}^2, \hat{\mathbf{S}}^3, \mathbf{E}^*$)))
\STATE $\mathbf{C} \leftarrow$ LN($\mathbf{C}$ + FFN($\mathbf{C}$))
\STATE // Step 4: Change caption prediction
\STATE $\mathbf{P} \leftarrow$ Softmax(Linear($\mathbf{C}$))
\STATE Predict caption words $y$ from the vocabulary using probability $\mathbf{P}$
\end{algorithmic}
\end{algorithm}
\subsection{Loss Function}
Most RSICC methods train caption generation networks using cross-entropy loss \cite{cai2023interactive, hoxha2022change, liu2023progressive}, and our model is trained using the same loss. Given a target ground-truth caption sequence $y_{1:T}^*$ and a captioning model with parameters $\theta$, the loss function minimizes the sum of the negative log-likelihood of the correctly predicted words at each step defined as follows:
\begin{equation}
L_{\text{CE}} = \sum_{t=1}^{T} -\log (p_\theta(y_t^* \mid y_{1:t-1}^*)),
\end{equation}
where $y_t^*$ denotes a one-hot vector for the $t$-th word in a ground truth sentence of length $T$. The model is trained to predict the target ground truth word $y_t^*$ conditioned on the previous words $y_{1:t-1}^*$.
\section{Experiments}\label{sec:experiments}
In this section, we first introduce the two public datasets used for our experiments, namely LEVIR-CC \cite{liu2022remote} and Dubai-CC \cite{hoxha2022change}. We then describe the evaluation metrics, implementation details, and baseline methods. Subsequently, we present comparisons with state-of-the-art methods. Finally, ablation studies are conducted to analyze the importance and contributions of each component.
\subsection{Datasets}
\subsubsection{LEVIR-CC Dataset}
The LEVIR-CC dataset is derived from the LEVIR building change detection dataset and consists of 10,077 bi-temporal image pairs, partitioned into 6,815 pairs for training, 1,333 pairs for validation, and 1,929 pairs for testing, each with a spatial resolution of $256 \times 256$ pixels. Each image pair is annotated to indicate whether it contains a change, with 5,038 changed pairs and 5,039 unchanged pairs, and accompanied by five sentence-level textual descriptions characterizing the changes. Following prior works \cite{liu2022remote, cai2023interactive}, the vocabulary used in our experiments includes only words that appear at least five times in the annotated sentences.
\subsubsection{Dubai-CC Dataset}
The Dubai-CC dataset provides a comprehensive description of urbanization changes in the Dubai region over a 10-year period. It consists of 500 bi-temporal image pairs with an original spatial resolution of $50 \times 50$ pixels. The dataset is divided into 300 image pairs for training, 50 for validation, and 150 for testing. Each image pair is manually annotated with five sentence-level descriptions detailing the nature of the observed changes. Following previous work \cite{li2024inter}, all original images are upsampled to a resolution of $256 \times 256$ pixels prior to being fed into the model.
\subsection{Evaluation Metrics}
To comprehensively evaluate the performance of our proposed model and the ablated models, we adopt a set of widely used evaluation metrics for RSICC: BLEU-n \cite{papineni2002bleu}, METEOR \cite{banerjee2005meteor}, ROUGE\_L \cite{chin2004rouge}, and CIDEr \cite{vedantam2015cider}. The BLEU-n metric includes BLEU-1-4, which measure the n-gram similarity between generated sentences and ground truth sentences. METEOR computes the harmonic mean of precision and recall of single words and incorporates a penalty factor to consider the fluency of generated sentences. ROUGE\_L measures the similarity of the longest common subsequence between generated sentences and ground truth sentences. CIDEr treats each sentence as a document and represents it in the form of term frequency--inverse document frequency (TF-IDF) vectors, with the score obtained by computing their cosine similarity. Using the comprehensive suite of evaluation metrics, we provide a thorough assessment of the change captioning capabilities. Higher scores across these metrics indicate more accurate generated captions.
\subsection{Implementation Details}
In our experiments, the proposed network is implemented in PyTorch and trained on NVIDIA H100 GPUs. The ResNet-101 backbone \cite{he2016deep} is initialized with pre-trained weights on ImageNet \cite{deng2009imagenet}. The initial learning rate is set to $1 \times 10^{-4}$ and decays by a factor of 0.7 when the sum of BLEU-4, ROUGE\_L, METEOR, and CIDEr scores shows no improvement over three consecutive epochs. Training is stopped when such aggregated metric fails to improve over 30 epochs. Model is optimized using the Adam optimizer \cite{kingma2014adam}. The dropout ratio is set to 0.5 to mitigate overfitting. In the ResNet-101 backbone, the parameters of layers up to and including stage 1 are frozen. Only stage 2 and 3 are fine-tuned during training. Stage 4 is discarded in our network. The remaining parameters in our network are fully optimized. The dimensions $C$, $d$, and $D$ are all set to 512. The change-aware encoder and the captioning decoder each use a single layer. Following previous works \cite{cai2023interactive, li2024inter}, the beam search size is set to 3 during inference. The batch size is set to 16. We apply data augmentation to the training set to enhance model generalization. Specifically, we randomly perform three types of flips, namely horizontal, vertical, and diagonal, and four types of rotations, namely 0\degree, 90\degree, 180\degree, and 270\degree, on each image pair. In addition, random color jitter is applied to both images, with brightness, contrast, and saturation factors sampled from $\pm 0.2$, and hue set to $0.0$. To ensure consistency, both images in each pair undergo identical transformations. Note that the total parameter count of our model differs slightly across datasets (39.99M on LEVIR-CC and 39.85M on Dubai-CC), as the word embedding and output projection layers scale with the size of each dataset's vocabulary.
\subsection{Baselines}
We evaluate our model by comparing it with several state-of-the-art methods, namely, Chg2Cap \cite{chang2023changes}, ICT-Net \cite{cai2023interactive}, TISDNet \cite{li2024inter}, KCFI \cite{yang2025enhancing}, CD4C \cite{li2025cd4c}, RSICCformer\cite{liu2022remote}, Prompt-CC \cite{liu2023decoupling}, TACC \cite{hang2025text}, SparseFocus \cite{sun2024lightweight}, SFEN \cite{zhang2024scale}, RSCaMa \cite{liu2024rscama}, DACC \cite{li2024detection}, SEN \cite{zhou2024single}, Change3D \cite{zhu2025change3d}, Semantic-CC \cite{zhu2024semantic}, MAF-Net \cite{chen2024multi}, MADiffCC \cite{yang2024remote}, SGD-RSCCN \cite{sun2025scene}, MFRNet \cite{xu2024mfrnet}, and DAE \cite{xian2025dynamic}. All of these results are taken directly from the original papers.
\begin{table}[t]
  \caption{Performance comparison on the LEVIR-CC dataset.
Results are reported in \%. \textbf{Bold} values represent the best results and \underline{underlined} values represent the second best. Parameter counts and FLOPs are the values reported in the original papers, with `-' indicating that the value is not reported. Following common practice, FLOPs are counted with one multiply--accumulate operation as one FLOP (i.e., MACs).}
  \label{tab:levir}
  \centering
  \resizebox{\linewidth}{!}{
  \begin{tabular}{@{}cccccccccc@{}}
    \toprule
    Method & BLEU-1 & BLEU-2 & BLEU-3 & BLEU-4 & METEOR & ROUGE\_L & CIDEr & Para (M) & FLOPs (G) \\
    \midrule
    Chg2Cap & 86.14 & 78.08 & 70.66 & 64.39 & 40.03 & 75.12 & 136.61 & 285.5 & - \\
    ICT-Net & 86.06 & 78.12 & 71.45 & 66.12 & 40.51 & 75.21 & 138.36 & 96.40 & - \\
    TISDNet & 86.79 & 79.35 & \underline{73.10} & \underline{67.93} & \textbf{41.60} & 76.64 & \underline{143.71} & 91.40 & - \\
    KCFI & 86.34 & 77.31 & 70.89 & 65.30 & 39.42 & 75.47 & 138.25 & 309.55 & - \\
    CD4C & 86.74 & 78.71 & 71.48 & 65.41 & 40.56 & 75.72 & 138.00 & 80.24 & - \\
    RSICCformer & 84.72 & 76.27 & 68.87 & 62.77 & 39.61 & 74.12 & 134.12 & 81.51 & - \\
    Prompt-CC & 83.66 & 75.73 & 69.10 & 63.54 & 38.82 & 73.72 & 136.44 & 408.58 & - \\
    TACC & 85.49 & 77.41 & 70.52 & 64.62 & 40.07 & 74.96 & 137.17 & 94.14 & - \\
    SparseFocus & 84.56 & 75.87 & 68.64 & 62.87 & 39.93 & 74.69 & 137.05 & 647.00 & 21.47 \\
    SFEN & 85.20 & 78.91 & 70.96 & 64.67 & 40.12 & 75.22 & 136.47 & 243.19 & - \\
    RSCaMa & 85.79 & 77.99 & 71.04 & 65.24 & 39.91 & 75.24 & 136.56 & 176.90 & 13.03 \\
    DACC & 87.26 & 79.16 & 71.74 & 65.56 & 40.65 & 75.97 & 139.99 & 73.0 & - \\
    DAE & \underline{87.40} & \textbf{79.99} & \textbf{73.71} & \textbf{68.52} & \underline{41.40} & 76.53 & \textbf{144.01} & 70.88 & - \\
    \midrule
    SEN & 85.10 & 77.05 & 70.01 & 64.09 & 39.59 & 74.57 & 136.02 & 39.90 & - \\
    Change3D & 85.81 & 77.81 & 70.57 & 64.38 & 40.03 & 75.12 & 138.29 & 5.05 & 2.39 \\
    \midrule
    Semantic-CC & \textbf{88.07} & \underline{79.68} & 71.47 & 64.51 & 40.58 & \textbf{77.76} & 138.51 & - & - \\
    MAF-Net & 85.15 & 78.00 & 71.89 & 66.78 & 39.72 & 74.43 & 134.97 & - & - \\
    MADiffCC & 86.28 & 77.50 & 71.09 & 66.93 & 40.16 & 75.37 & 138.61 & - & - \\
    SGD-RSCCN & 84.17 & 75.16 & 68.05 & 62.48 & 39.18 & 74.24 & 136.27 & - & - \\
    MFRNet & 85.86 & 78.29 & 72.13 & 67.21 & 40.22 & 75.19 & 139.56 & - & - \\
    \midrule
    LBTCap (\textbf{Ours}) & 86.61 & 78.77 & 71.75 & 65.94 & 40.96 & \underline{76.74} & 142.57 & 39.99 & 21.82 \\
    \bottomrule
  \end{tabular}
  }
\end{table}
\begin{table}[t]
  \caption{Performance comparison on the Dubai-CC dataset.
Results are reported in \%. \textbf{Bold} values represent the best results and \underline{underlined} values represent the second best.}
  \label{tab:dubai}
  \centering
  \resizebox{\linewidth}{!}{
  \begin{tabular}{@{}cccccccc@{}}
    \toprule
    Method & BLEU-1 & BLEU-2 & BLEU-3 & BLEU-4 & METEOR & ROUGE\_L & CIDEr \\
    \midrule
    Chg2Cap & 72.04 & 60.16 & 50.84 & 41.70 & 28.92 & 58.66 & 92.49 \\
    ICT-Net & 69.38 & 57.03 & 46.50 & 36.17 & 26.78 & 57.31 & 92.97 \\
    TISDNet & 71.81 & 60.60 & 50.88 & 41.37 & 29.67 & 63.24 & 109.47 \\
    CD4C & 74.13 & 63.11 & 53.13 & 43.55 & 30.27 & 60.25 & 108.66 \\
    TACC & 72.17 & 58.65 & 48.24 & 38.59 & 28.17 & 58.93 & 95.45 \\
    SparseFocus & 67.30 & 55.97 & 47.00 & 37.30 & 26.32 & 56.38 & 91.59 \\
    \midrule
    SEN & 70.95 & 57.28 & 45.81 & 36.25 & 26.62 & 55.95 & 91.77 \\
    Change3D & 72.25 & 58.68 & 47.13 & 36.80 & 27.06 & 56.04 & 86.19 \\
    \midrule
    MADiffCC & 73.18 & 61.36 & 52.25 & 45.41 & 30.85 & 60.56 & 96.47 \\
    SGD-RSCCN & \textbf{77.15} & \textbf{66.80} & \textbf{58.07} & \textbf{50.27} & \textbf{34.07} & 65.88 & \underline{118.42} \\
    DAE & \underline{76.08} & \underline{64.77} & \underline{54.95} & 44.76 & 32.38 & \textbf{68.06} & 113.94 \\
    \midrule
    LBTCap (\textbf{Ours}) & 75.20 & 64.00 & 54.57 & \underline{45.90} & \underline{33.25} & \underline{66.86} & \textbf{118.81} \\
    \bottomrule
  \end{tabular}
  }
\end{table}
\subsection{Evaluation Results and Analysis}
\subsubsection{Results on the LEVIR-CC Dataset}
We first present the comparative results on the LEVIR-CC dataset. The compared methods are divided into three groups: those with significantly more parameters than ours, those with a similar or smaller number of parameters, and those whose parameter sizes are unavailable. Table~\ref{tab:levir} reports the results. As shown in Table~\ref{tab:levir}, within the first group, our method achieves accuracy comparable to or better than most competitors (Chg2Cap, ICT-Net, KCFI, CD4C, RSICCformer, Prompt-CC, TACC, SparseFocus, SFEN, RSCaMa, and DACC) while using far fewer parameters.  Although TISDNet attains the highest BLEU scores, our method remains highly competitive while requiring less than half of its parameters (39.99M vs. 91.40M). Within the second group, which comprises methods with a similar or smaller number of parameters (SEN and Change3D), our method outperforms both across all seven metrics. Notably, although Change3D is far smaller (5.05M vs. 39.99M), our method is clearly more accurate. The recent DAE reports higher scores than ours on most metrics, but at roughly 1.8 times the model size (70.88M vs. 39.99M). Within the third group, whose parameter sizes are unavailable, our method attains the highest CIDEr, clearly surpassing SGD-RSCCN and achieving the best CIDEr among Semantic-CC, MAF-Net, MADiffCC, and MFRNet while remaining competitive on the remaining metrics. These results demonstrate that our method strikes a favorable balance between accuracy and model size, achieving strong performance with low computational cost.
\subsubsection{Results on the Dubai-CC Dataset}
We then present the comparative results on the Dubai-CC dataset, following the same three-group categorization, as summarized in Table~\ref{tab:dubai}. In the first group, consisting of methods with significantly more parameters than ours, our method outperforms all competing approaches across all seven metrics, including Chg2Cap, ICT-Net, TISDNet, CD4C, TACC, and SparseFocus. Within the second group, comprising methods with a similar or smaller number of parameters, such as SEN and Change3D, our method consistently achieves superior performance across all seven metrics. Within the third group, where parameter information is unavailable, our method outperforms MADiffCC across all seven metrics and surpasses SGD-RSCCN on two of the seven metrics. Notably, although the exact parameter size of SGD-RSCCN is not reported, it employs a full ResNet-101 backbone, which alone exceeds 45M parameters, excluding additional components such as the change-aware encoder and captioning decoder, indicating that its overall model size is substantially larger than ours. These results further demonstrate that our method provides an effective trade-off between accuracy and model size.

We further compare our approach with the two competing methods that outperform it on each individual dataset, considering both the LEVIR-CC and Dubai-CC datasets. The cross-dataset comparison shows that, although TISDNet performs better on LEVIR-CC, our method substantially surpasses it on Dubai-CC. Similarly, while SGD-RSCCN achieves higher performance on Dubai-CC, our method clearly outperforms it on LEVIR-CC. In contrast, neither competitor remains strong on both datasets. Overall, at a markedly smaller model size, our method attains a more consistent accuracy across the two datasets than these larger competitors, delivering a favorable and stable accuracy--efficiency trade-off.

We further observe that our method is relatively stronger on CIDEr, ROUGE\_L, and METEOR than on the higher-order BLEU metrics, and we attribute this to what these metrics emphasize. BLEU-$n$ measures the precision of exact $n$-gram matches and is therefore sensitive to the specific wording and word order of a caption, particularly for large $n$. CIDEr, in contrast, weights $n$-grams by TF-IDF, which down-weights the frequent template tokens that dominate RSICC captions and rewards the informative content words that describe the actual change, while METEOR and ROUGE\_L further credit stem/synonym and longest-common-subsequence matches rather than exact phrasing. Since our bilateral change-aware encoder is designed to localize and describe the changed content, it reliably produces the correct change-related words but may phrase them differently from a particular reference, which raises CIDEr/ROUGE\_L/METEOR while slightly lowering higher-order BLEU. As CIDEr is generally regarded as correlating better with human judgment of change-description quality, we treat it as the more indicative metric for this task.
\subsubsection{Comparison of Parameters and Speed}
We regard an RSICC model as operating in real time when it processes incoming image pairs faster than they are acquired in operational monitoring pipelines, whose ingestion rate is typically only a few image pairs per second. Following common practice, we quantify this capability by the number of image pairs processed per second (FPS), where a higher FPS indicates stronger real-time capability. Across all tested GPUs, LBTCap runs at 18.94--62.26 FPS, which exceeds such acquisition rates by a wide margin and therefore meets the real-time requirement. To assess this, we measured the inference speed of our method and compared it with state-of-the-art methods. The results are summarized in Table~\ref{tab:para}. As shown in Table~\ref{tab:para}, our method achieves substantially higher FPS compared to RSICCformer, ICT-Net, and TISDNet on NVIDIA GeForce GTX 1080 Ti (18.94 vs. 12.40), corresponding to a 52.74\% increase. On NVIDIA GeForce RTX 4090, it outperforms Prompt-CC, FST-Net, and TACC (62.26 vs. 31.35). Similarly, on NVIDIA GeForce RTX 3090, our method surpasses SEN and Change3D (39.28 vs. 23.70). Notably, although Change3D has significantly fewer parameters, its FPS is relatively low, possibly due to its complex video encoder. Finally, on NVIDIA A100, our method achieves higher FPS than Chg2Cap (27.21 vs. 11.98). Our method's FPS follows the order RTX 4090 $>$ RTX 3090 $>$ A100 $>$ GTX 1080 Ti, consistent with the overall capability of these GPUs for our latency-bound workload. Since the speed is measured with a batch size of $1$, i.e., one image pair is processed at a time with sequential caption decoding, the A100's large-batch throughput advantage cannot be exploited. Among the same-generation Ampere cards, the higher-clocked RTX 3090 therefore slightly exceeds the A100. These results demonstrate that our approach provides substantially improved real-time performance over prior methods. In addition, the Mamba-based RSCaMa involves 176.90M parameters and 13.03G FLOPs, as reported in its original paper~\cite{liu2024rscama}. For LBTCap, the FLOP count measured at the $256\times256$ input resolution is 21.82G (counted as MACs, following the same convention as the other reported values), which we obtain with ptflops and cross-check with thop. It is comparable to efficient baselines such as SparseFocus (21.47G) despite LBTCap's far smaller parameter count, and, consistent with Table~\ref{tab:params}, most of it stems from the ResNet-101 backbone rather than the lightweight bilateral encoder. For transparency, Table~\ref{tab:params} reports the per-component parameter breakdown of our 39.99M learnable parameters. Notably, the change-aware encoder, which is the core of our method, adds only 2.78M parameters, confirming that most of the model size stems from the shared backbone rather than from the proposed component.

\begin{table}[t]
  \caption{Per-component parameter breakdown of LBTCap (learnable parameters on LEVIR-CC).}
  \label{tab:params}
  \centering
  \begin{tabular}{@{}lc@{}}
    \toprule
    Component & Parameters (M) \\
    \midrule
    Feature extractor (ResNet-101) & 27.54 \\
    Change-aware encoder & 2.78 \\
    Captioning decoder & 9.67 \\
    \midrule
    Total & 39.99 \\
    \bottomrule
  \end{tabular}
\end{table}
\begin{table}[t]
\footnotesize
  \caption{Comparison with state-of-the-art methods in terms of parameter count and inference speed (image pairs per second). The FPS values of the competing methods are taken from their original papers on the GPU reported therein, except for Change3D, which does not report speed and is measured by us. The FPS of LBTCap is measured by us with a batch size of 1.}
  \label{tab:para}
  \centering
  \begin{tabular}{@{}cccc@{}}
    \toprule
    Method & Parameter (M) & GPU & FPS \\
    \midrule
    RSICCformer & 81.51 & 1080Ti & 12.40 \\
    ICT-Net & 96.40 & 1080Ti & 10.44 \\
    TISDNet & 91.40 & 1080Ti & 11.61 \\
    LBTCap (\textbf{Ours}) & 39.85 & 1080Ti & \textbf{18.94} \\
    \midrule
    Prompt-CC & 408.58 & 4090 & 9.94 \\
    FST-Net \cite{zou2025frequency} & 111.76 & 4090 & 31.35 \\
    TACC & 94.14 & 4090 & 0.94 \\
    LBTCap (\textbf{Ours}) & 39.85 & 4090 & \textbf{62.26} \\
    \midrule
    SEN & 39.90 & 3090 & 23.70 \\
    Change3D & 5.05 & 3090 & 9.47 \\
    LBTCap (\textbf{Ours}) & 39.85 & 3090 & \textbf{39.28} \\
    \midrule
    Chg2Cap & 285.5 & A100 & 11.98 \\
    LBTCap (\textbf{Ours}) & 39.85 & A100 & \textbf{27.21} \\
    \bottomrule
  \end{tabular}
\end{table}
\begin{figure*}[t]
  \centering
  \subfloat[Example results from the LEVIR-CC dataset.]{
     \includegraphics[width=\linewidth]{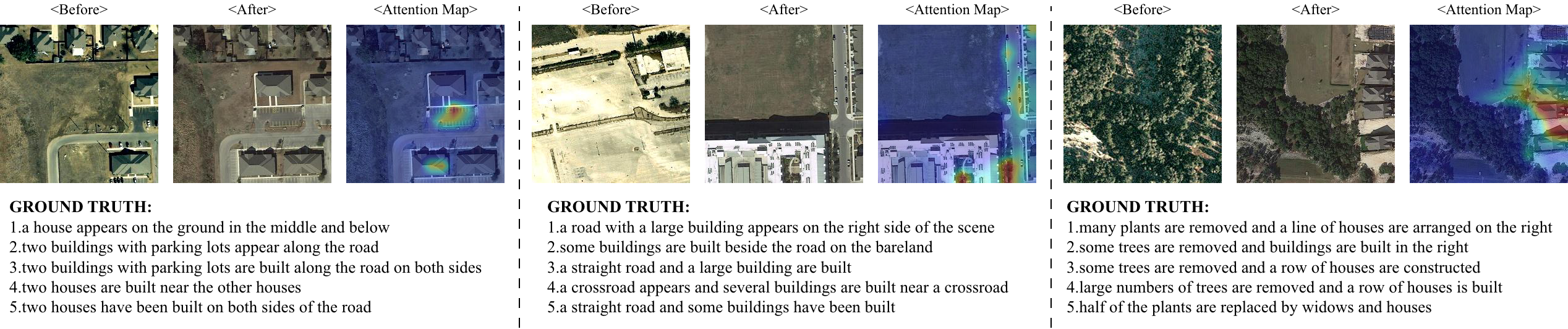}
     \label{fig:LEVIR}
  }
  \hfill
  \subfloat[Example results from the Dubai-CC dataset.]{
     \includegraphics[width=\linewidth]{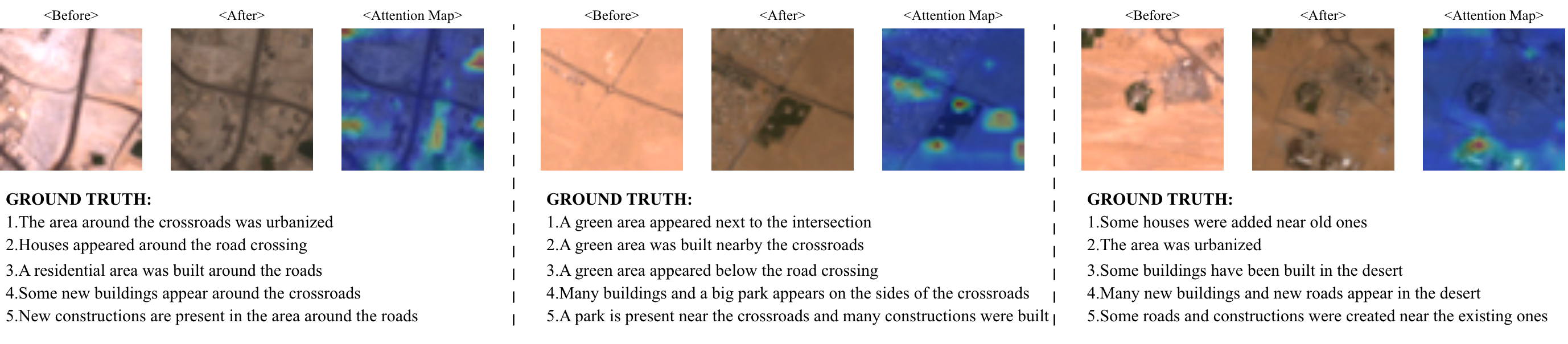}
     \label{fig:Dubai}
  }
  \caption{Attention maps of the post-change branch (the softmax$(Q_2K_2^{T})$ map on the `after' RSI) generated by LBTCap on: (a) the LEVIR-CC dataset, and, (b) the Dubai-CC dataset.}
  \label{fig:attentionmap}
\end{figure*}
\begin{figure}[t]
  \centering
   \includegraphics[width=\linewidth]{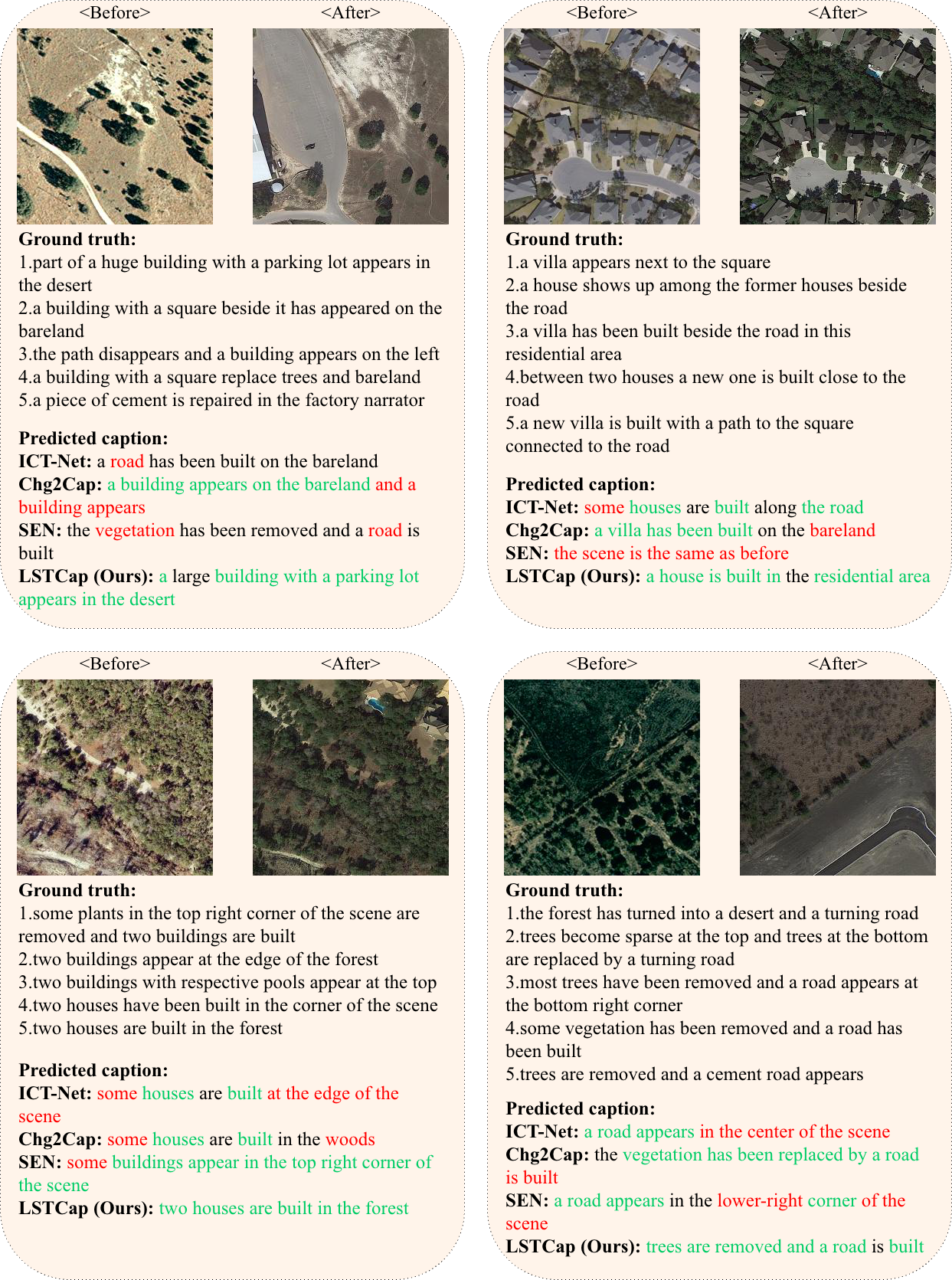}
   \caption{Captioning results compared with other methods. The red font represents wrong predictions and the green font represents correct predictions.}
   \label{fig:results}
\end{figure}

To further illustrate the accuracy--efficiency trade-off of our method, we present a comparative plot of both accuracy and model size across different approaches. We evaluate our method against SparseFocus~\cite{sun2024lightweight}, RSICCformer~\cite{liu2022remote}, TISDNet~\cite{li2024inter}, Chg2Cap~\cite{chang2023changes}, SEN~\cite{zhou2024single}, ICT-Net~\cite{cai2023interactive}, CD4C~\cite{li2025cd4c}, Prompt-CC~\cite{liu2023decoupling}, KCFI~\cite{yang2025enhancing}, RSCaMa~\cite{liu2024rscama}, TACC~\cite{hang2025text}, SFEN~\cite{zhang2024scale}, DACC~\cite{li2024detection}, Change3D~\cite{zhu2025change3d}, and DAE~\cite{xian2025dynamic}, using parameter size on the horizontal axis and the BLEU-4 score (in percentage) on the LEVIR-CC dataset on the vertical axis. In this setting, we consider the BLEU-4 metric on the LEVIR-CC dataset. As illustrated in Fig.~\ref{fig:acc_param}, our model achieves a BLEU-4 score of 65.94\% with only 39.99M parameters. Although a few methods attain higher BLEU-4 scores, they rely on substantially larger models (e.g., TISDNet reaches 67.93\% with 91.40M parameters). Among methods with a comparable or smaller number of parameters, our model attains the highest BLEU-4, demonstrating that a compact and efficient design can still deliver competitive accuracy.
\begin{table*}[t]
\caption{Ablation study of multi-scale mechanism on LEVIR-CC and Dubai-CC datasets. Results are reported in \%.}
\label{tab:multiscale}
\centering
\resizebox{1.0\linewidth}{!}{
\begin{tabular}{@{}cccccccccc@{}}
\toprule
   Dataset & Multi-scale Choice & BLEU-1 & BLEU-2 & BLEU-3 & BLEU-4 & METEOR & ROUGE\_L & CIDEr & Parameter (M) \\
    \midrule
    \multirow{2}{*}{LEVIR-CC} & Three stages & 85.33 & 77.81 & 71.45 & \textbf{66.51} & 40.39 & 75.09 & 138.18 & 56.79 \\
    & Two stages (\textbf{Ours}) & \textbf{86.61} & \textbf{78.77} & \textbf{71.75} & 65.94 & \textbf{40.96} & \textbf{76.74} & \textbf{142.57} & 39.99 \\
    \midrule
    \multirow{2}{*}{Dubai-CC} & Three stages & 74.19 & 62.61 & 53.35 & 45.04 & 31.85 & \textbf{67.14} & 118.37 & 56.65 \\
    & Two stages (\textbf{Ours}) & \textbf{75.20} & \textbf{64.00} & \textbf{54.57} & \textbf{45.90} & \textbf{33.25} & 66.86 & \textbf{118.81} & 39.85 \\
\bottomrule
\end{tabular}
}
\end{table*}
\subsection{Qualitative Comparison}
To verify the effectiveness of LBTCap, we present visualization examples of the attention maps generated by our model on the LEVIR-CC and Dubai-CC datasets in Fig.~\ref{fig:attentionmap}. As shown in Fig.~\ref{fig:LEVIR}, the attention maps accurately highlight the changed objects, aligning well with the ground truth and focusing on regions of visual change, thereby demonstrating the model's capability to effectively capture changes between image pairs. Furthermore, as illustrated in Fig.~\ref{fig:Dubai}, despite the lower resolution and the increased difficulty in identifying changed regions, the attention maps still localize these areas with relatively high precision.

Furthermore, we present several examples to demonstrate the accuracy of the captions generated by our model. The visualization in Fig.~\ref{fig:results} shows the RSICC results produced by our method and several baselines on the LEVIR-CC dataset, along with the corresponding ground truth annotations for comparison. Green text indicates correct descriptions, while red text denotes incorrect ones. Across these examples, our method generally produces more accurate captions by identifying the key changes in the scenes, whereas the compared models more often miss critical details, introduce incorrect information, or generate incoherent sentences. Specifically, in the top-left example, ICT-Net and SEN produce descriptions containing non-existent content, while Chg2Cap generates a repetitive and incoherent sentence. In contrast, our method produces an accurate caption. In the top-right example, ICT-Net predicts incorrect information, Chg2Cap introduces irrelevant content, and SEN incorrectly classifies the scene as unchanged, whereas our method generates a more accurate description. In the bottom-left example, ICT-Net, Chg2Cap, and SEN all produce inaccurate captions, while our method yields a correct sentence. In the bottom-right example, ICT-Net and SEN generate incorrect descriptions, and Chg2Cap produces an incoherent sentence, whereas our method again produces a correct caption. Overall, these examples demonstrate that our model effectively focuses on critical change regions, resulting in more precise and coherent captions that are closely aligned with the ground truth annotations, thereby validating the effectiveness of the proposed approach.
\begin{table}[t]
\caption{Ablation study of change-aware function on the LEVIR-CC and Dubai-CC datasets. Results are reported in \%.}
\label{tab:difffunc}
\centering
\resizebox{1.0\linewidth}{!}{
\begin{tabular}{@{}ccccccccc@{}}
\toprule
   Dataset & Change-aware Function & BLEU-1 & BLEU-2 & BLEU-3 & BLEU-4 & METEOR & ROUGE\_L & CIDEr \\
    \midrule
    \multirow{2}{*}{LEVIR-CC} & DiffAttn \cite{yedifferential} & 86.42 & 78.68 & \textbf{72.12} & \textbf{66.86} & \textbf{41.31} & 76.22 & 140.58 \\
    & \textbf{Ours} & \textbf{86.61} & \textbf{78.77} & 71.75 & 65.94 & 40.96 & \textbf{76.74} & \textbf{142.57} \\
    \midrule
    \multirow{2}{*}{Dubai-CC} & DiffAttn & 71.89 & 60.32 & 50.71 & 41.10 & 31.14 & 63.69 & 108.89 \\
    & \textbf{Ours} & \textbf{75.20} & \textbf{64.00} & \textbf{54.57} & \textbf{45.90} & \textbf{33.25} & \textbf{66.86} & \textbf{118.81} \\
\bottomrule
\end{tabular}
}
\end{table}

\begin{table*}[t]
\caption{Ablation study of key--value heads on the LEVIR-CC and Dubai-CC datasets. Results are reported in \%. `FED' denotes feature embedding dimension, `KV-Heads' indicates the number of key--value heads, and `Heads' refers to the number of query heads.}
\label{tab:kvheads}
\resizebox{1.0\textwidth}{!}{
\begin{tabular}{@{}cccccccccccc@{}}
\toprule
   Dataset & FED & Heads & KV-Heads & BLEU-1 & BLEU-2 & BLEU-3 & BLEU-4 & METEOR & ROUGE\_L & CIDEr & Parameter (M) \\
    \midrule
    \multirow{3}{*}{LEVIR-CC} & 512 & 8 & 8 & \textbf{86.90} & \textbf{78.99} & \textbf{71.96} & \textbf{66.08} & \textbf{40.99} & 76.11 & 140.37 & 40.38 \\
    & 512 & 8 & 4 & 86.61 & 78.77 & 71.75 & 65.94 & 40.96 & \textbf{76.74} & \textbf{142.57} & 39.99 \\
    & 512 & 8 & 2 & 85.52 & 77.60 & 70.86 & 65.52 & 40.94 & 75.37 & 138.64 & 39.79 \\
    \midrule
    \multirow{3}{*}{Dubai-CC} & 512 & 8 & 8 & 73.28 & 62.96 & 54.38 & 45.20 & 31.68 & \textbf{67.26} & \textbf{121.35} & 40.24 \\
    & 512 & 8 & 4 & \textbf{75.20} & \textbf{64.00} & \textbf{54.57} & \textbf{45.90} & \textbf{33.25} & 66.86 & 118.81 & 39.85 \\
    & 512 & 8 & 2 & 71.71 & 61.88 & 53.19 & 44.83 & 31.69 & 64.54 & 112.94 & 39.65 \\
\bottomrule
\end{tabular}
}
\end{table*}
\subsection{Ablation Experiments}
To evaluate the contribution of each component in our model, we conduct comprehensive ablation studies.
\subsubsection{Effectiveness of our proposed multi-scale mechanism}
As discussed in Section~\ref{sec3.1}, our method relies on feature maps from stages 2 and 3 while discarding those from stage 4, which could potentially reduce accuracy. To evaluate the impact of the design choice, we compare the performance between the two settings. Specifically, we conduct additional experiments that incorporate feature maps from all three stages in the multi-scale feature extraction process. Table~\ref{tab:multiscale} presents the comparative results of these two settings on both datasets. As shown in Table~\ref{tab:multiscale}, on the LEVIR-CC dataset, the two-stage setting outperforms the three-stage variant on six out of seven metrics. Similarly, on the Dubai-CC dataset, the two-stage setting achieves better performance on six out of seven metrics. These results indicate that excluding the stage-4 features does not degrade performance. Moreover, the two-stage configuration significantly reduces the number of model parameters, from 56.79M to 39.99M and from 56.65M to 39.85M on the two datasets, respectively, demonstrating that our proposed multi-scale mechanism achieves a more favorable trade-off between performance and model complexity. We attribute this to the nature of the discarded stage-4 features: stage 4 produces low-resolution, highly abstract feature maps whose spatial detail is largely lost, which is detrimental to describing fine-grained and small-scale changes, and the extra parameters it introduces increase the risk of overfitting on the relatively small RSICC training sets rather than adding useful change cues. Retaining only stages 2 and 3 preserves the mid- and high-level spatial semantics that are most informative for change localization, so removing stage 4 improves accuracy and efficiency simultaneously.
\subsubsection{Impact of change-aware function}
We further investigate the effectiveness of the proposed change-aware function through a comprehensive ablation study. To this end, we replace our design with the original function proposed in \cite{yedifferential} (denoted as DiffAttn) and conduct controlled experiments under identical settings. Table~\ref{tab:difffunc} summarizes the comparative results of the two functions on the LEVIR-CC and Dubai-CC datasets. As shown in Table~\ref{tab:difffunc}, on the LEVIR-CC dataset the two functions perform comparably: our design achieves higher BLEU-1, BLEU-2, ROUGE\_L, and CIDEr, while DiffAttn is slightly better on BLEU-3, BLEU-4, and METEOR, with all differences remaining small. On the lower-resource Dubai-CC dataset, however, our function outperforms DiffAttn across all seven metrics by clear margins (e.g., 45.90 vs. 41.10 on BLEU-4 and 118.81 vs. 108.89 on CIDEr). We attribute this to the bilateral structure of our design, which treats both inputs through the same operation and thus yields a more stable change representation when training data are scarce. In contrast, the subtractive formulation of DiffAttn can bias the interaction toward one input. Overall, the proposed function is on par with DiffAttn on the data-rich LEVIR-CC dataset and offers a clear advantage in the low-resource Dubai-CC setting, which motivates our bilateral design.

To further understand this design, we inspect the converged value of $\lambda$ (with $\lambda_{\mathrm{init}}=0.2$ for the single encoder layer), as summarized in Table~\ref{tab:lambda}. Two points stand out. First, on both datasets $\lambda$ departs substantially from $0.5$, indicating that the model does not weight the two temporal branches equally but instead learns a dataset-specific balance. Second, the direction of this balance is opposite across datasets: on LEVIR-CC the combination emphasizes the post-change branch and assigns a negative weight to the pre-change one, while on Dubai-CC it emphasizes the pre-change branch and assigns a negative weight to the post-change one. This is precisely the flexibility that our bilateral formulation adds over the differential attention of~\cite{yedifferential}: the original form always keeps a unit weight on the first map and subtracts a bounded amount of the second, and therefore cannot reverse which branch is suppressed, whereas our form allows either branch to be down-weighted as the data require. The fact that the two datasets converge to opposite balances thus supports the practical value of this added flexibility, beyond its structural motivation.

\begin{table}[t]
  \caption{Converged balancing weight $\lambda$ of the change-aware function on each dataset ($\lambda_{\mathrm{init}}=0.2$). The two attention maps are combined as $(1-\lambda)\,\mathrm{softmax}(A_1)+\lambda\,\mathrm{softmax}(A_2)$, where $A_1$ and $A_2$ are the pre- and post-change maps.}
  \label{tab:lambda}
  \centering
  \resizebox{\linewidth}{!}{
  \begin{tabular}{@{}lccc@{}}
    \toprule
    Dataset & $\lambda$ & Pre-change weight $(1-\lambda)$ & Emphasized branch \\
    \midrule
    LEVIR-CC & $1.68$ & $-0.68$ & post-change \\
    Dubai-CC & $-0.45$ & $1.45$ & pre-change \\
    \bottomrule
  \end{tabular}
  }
\end{table}
\subsubsection{Influence of GQA}
We further investigate the influence of GQA, which we employ in the multi-head attention mechanism (Eq.~\eqref{eqn:gqa}) where multiple query heads share the same key and value projections. To this end, we conduct ablation studies to examine how variations in hyperparameters affect both model size and accuracy. The key hyperparameters include the feature embedding dimension, the number of query heads, and the number of key--value heads. In this study, we fix the feature embedding dimension at 512 and set the number of query heads to 8, as larger values are not applicable. Increasing the feature embedding dimension would significantly increase the model size. We then vary the number of key--value heads to compare different settings, adopting three configurations---2, 4, and 8---all of which evenly divide 8. Table~\ref{tab:kvheads} presents the comparative results of these configurations along with their corresponding parameter sizes on the LEVIR-CC and Dubai-CC datasets.

As shown in Table~\ref{tab:kvheads}, although the model with only 2 key--value heads has the smallest size, its performance consistently lags behind the models with 4 and 8 key--value heads. On the LEVIR-CC dataset, the model with 4 key--value heads performs slightly worse than the 8 key--value heads model on BLEU-n and METEOR, yet surpasses it on ROUGE\_L and achieves a substantially higher score on CIDEr. On the Dubai-CC dataset, the 4 key--value heads model outperforms the 8 key--value heads model on BLEU-n and METEOR, but underperforms on ROUGE\_L and substantially on CIDEr. Overall, considering both datasets, the two configurations achieve comparable performance, while the 4 key--value heads model reduces the parameter count by 0.39M. Considering the trade-off between model size and performance, we adopt the 4 key--value-heads configuration in our final model. The metric-level fluctuations across datasets reflect their different scales: the larger and more diverse LEVIR-CC can exploit the extra capacity of eight key--value heads on some metrics, whereas the much smaller Dubai-CC (300 training pairs) is more prone to overfitting and thus generalizes better with fewer heads. The residual differences are small and lie within the run-to-run variance expected on such a small test set, which is why the 4-head configuration remains the most balanced choice overall.
\section{Limitations}\label{sec:limitations}
We state the scope boundaries of our study explicitly. First, the evaluation covers two public RSICC benchmarks, LEVIR-CC and Dubai-CC; broader validation on additional datasets would further probe generalization. Second, most of the parameter and FLOPs savings come from the truncated ResNet-101 backbone and grouped-query attention, whereas the bilateral encoder itself contributes 2.78M parameters, so the encoder is best viewed as an accuracy-oriented, backbone-agnostic module rather than the sole source of efficiency. Third, the advantage of the bilateral formulation over differential attention is most pronounced in the low-resource setting, while on the data-rich LEVIR-CC the two perform on par. These boundaries delineate where the method is currently validated and where its benefits are strongest.

\section{Conclusion}\label{sec:conclu}
In this work, we propose an RSICC framework, LBTCap, designed to address the challenges of real-time RSICC deployment. At its core, LBTCap introduces a new bilateral attention mechanism for paired-input change modeling. It generates queries and keys separately from the two temporal images and forms the value from their concatenation, enabling simultaneous processing of paired inputs, and it combines the two attention maps through a learnable bilateral weighting that treats pre- and post-change features consistently. The mechanism performs on par with differential attention when training data are abundant and yields clear gains in the low-resource setting. We conduct extensive experiments comparing our method with state-of-the-art approaches and perform comprehensive ablation studies. The results demonstrate LBTCap's real-time capability, its favorable trade-off between accuracy and model size, and the effectiveness of the individual components in our model.

\section*{CRediT authorship contribution statement}
\textbf{Licheng Zhang:} Conceptualization, Methodology, Software, Validation, Writing -- original draft. \textbf{Siew-Kei Lam:} Supervision, Writing -- review \& editing. \textbf{Naveed Akhtar:} Supervision, Writing -- review \& editing.

\section*{Declaration of Competing Interest}
The authors declare that they have no known competing financial interests or personal relationships that could have appeared to influence the work reported in this paper.

\section*{Code and Data Availability}
The LEVIR-CC and Dubai-CC datasets analyzed in this study are publicly available benchmark datasets and are cited in the text. The code will be made available from the corresponding author upon reasonable request.

\bibliographystyle{elsarticle-num-names}
\bibliography{main}
\end{document}